\documentclass{article}
\PassOptionsToPackage{numbers, compress}{natbib}
\usepackage[final]{neurips_2026}
\makeatletter
\providecommand{\@trackname}{}
\makeatother

\usepackage[utf8]{inputenc}
\usepackage[T1]{fontenc}
\usepackage{url}
\usepackage{booktabs}
\usepackage{amsfonts}
\usepackage{nicefrac}
\usepackage{microtype}
\usepackage{xcolor}
\usepackage[table]{xcolor}
\usepackage{graphicx}
\usepackage{amsmath}
\usepackage{amssymb}
\usepackage{multirow}
\usepackage{threeparttable}
\usepackage{makecell}
\usepackage[colorlinks=true, linkcolor=blue, citecolor=blue, urlcolor=blue]{hyperref}
\usepackage{wrapfig}
\usepackage{booktabs}   
\usepackage{subcaption} 
\usepackage{amssymb}    

\title{FrequencyBooster: Full-Frequency Modeling for High-Fidelity Pixel Diffusion}

\author{
    \textbf{Lichen Ma}\thanks{Equal contribution} ,
    \textbf{Zipeng Guo}\footnotemark[1] ,
    \textbf{Yu He}\footnotemark[1] ,
    \textbf{Xiaolong Fu} ,
    \textbf{Luohang Liu} ,
    \textbf{Jingling Fu} \\
    \textbf{Junshi Huang}\thanks{Corresponding Author.} ,
    \textbf{Yan Li} \\
    {\tt\small \{malichen2020, junshi.huang\}@gmail.com}
}

\begin{document}

\maketitle

\begin{abstract}
To circumvent the inherent fidelity bottlenecks and optimization misalignment of VAE-based latent diffusion, pixel-space diffusion models have emerged as a compelling end-to-end paradigm. However, existing pixel diffusion models often struggle to balance computational efficiency with the preservation of high-frequency details. They frequently resort to patch-based compression or restricted local decoding, leading to a "spectral compromise" where high-frequency and fine-grained pixel information are suppressed. To address these challenges, we propose \textbf{FrequencyBooster}, a novel framework designed to empower pixel diffusion with full-frequency modeling capabilities without prohibitive overhead. 
The core of our method is a high-capacity decoder that specializes in extracting exhaustive high-frequency details and low-frequency semantics, the latter of which is derived from a Diffusion Transformer (DiT) backbone.
Unlike prior works that sacrifice global context for local refinement, FrequencyBooster leverages high-dimensional feature representations to maintain global structural integrity while achieving superior pixel-level precision. Extensive experiments on ImageNet demonstrate the effectiveness of our approach: our model achieves a state-of-the-art FID of \textbf{1.60} at $256 \times 256$ resolution within only 320 epochs. Furthermore, at $512 \times 512$ resolution, FrequencyBooster attains an FID of \textbf{1.69}, significantly outperforming existing pixel-space and latent-space generative models.
\end{abstract}

\section{Introduction}

\begin{figure*}[h]
    \centering
    \includegraphics[width=0.98\textwidth]{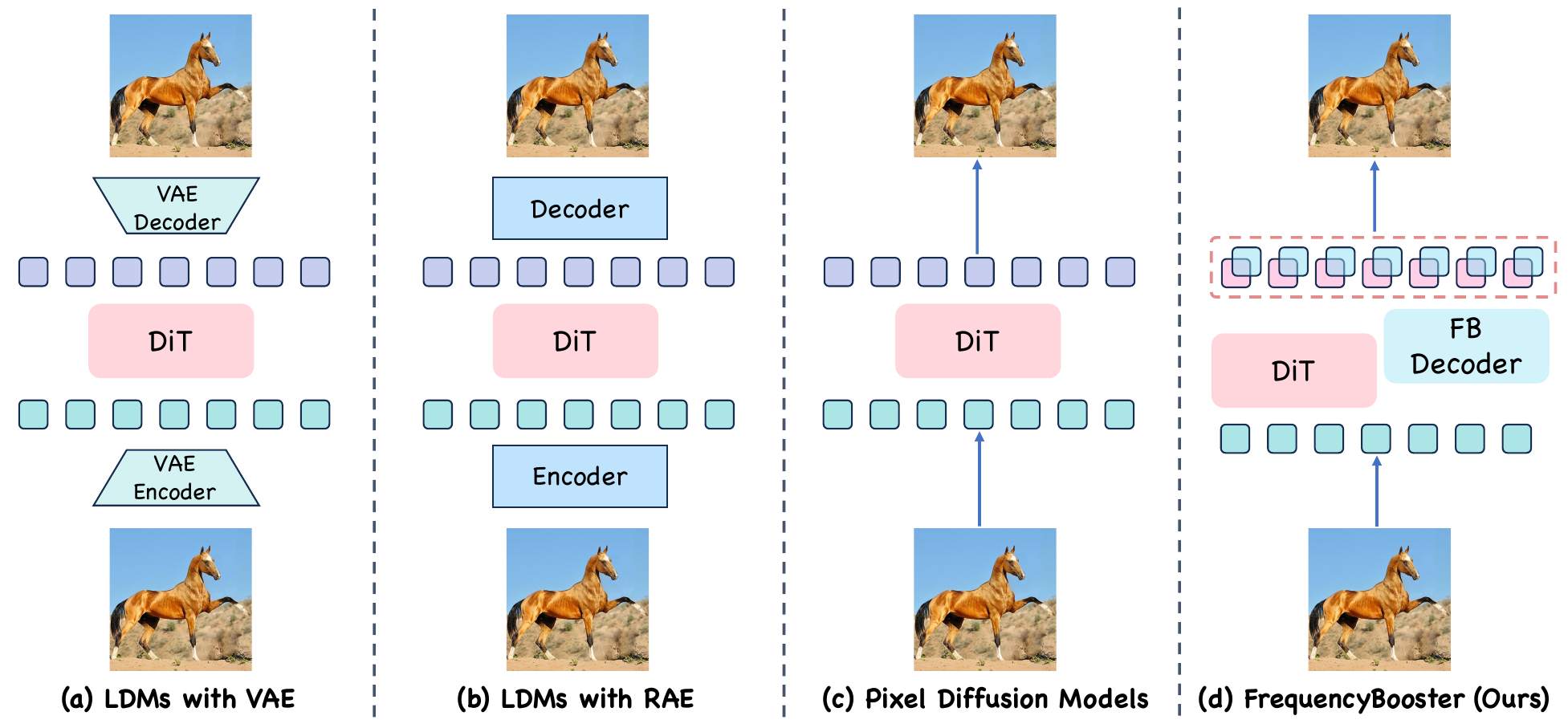}
    \caption{Architectures of (a) vanilla Latent Diffusion Models with VAE, (b) Representation for Generation Models (RAE), (c) vanilla Pixel-space Diffusion Models, and (d) our proposed method.}
    \label{fig:main}
    \vspace{-15pt}
\end{figure*}

Latent Diffusion Models(LDMs)\cite{rombach2022high} have fundamentally reshaped the landscape of visual content generation. Research in this field typically bifurcates into two primary trajectories: latent diffusion\cite{wang2025ddt, peebles2023scalable, ma2024sit, yu2025repa, zheng2023fast} and pixel diffusion\cite{tian2024u, ma2025deco, yu2025pixeldit}. 
While Latent Diffusion Models (LDMs) are the preferred standard for Diffusion Transformers due to their efficiency, the coupling of VAEs and diffusion processes introduces critical limitations. First, objective misalignment: pre-trained autoencoders use reconstruction criteria that diverge from downstream generative goals, causing optimization shifts. Second, precision bottlenecks: lossy compression triggers irreversible information decay, imposing a "fidelity ceiling" tied to the autoencoder's limits. To bypass these constraints, we propose pixel-space diffusion. By modeling raw pixels directly, our approach eliminates the performance degradation inherent in intermediate representations.

Pixel-space diffusion bypasses the inherent limitations of VAEs by modeling raw pixels directly in an end-to-end manner. 
Although the optimal approach would entail direct pixel-level prediction, such a design typically leads to prohibitive memory overhead, frequently resulting in out-of-memory (OOM) errors. 
Consequently, many existing pixel diffusion models \cite{li2025jit, ma2025deco, yu2025pixeldit} resort to patch-based tokenization to strike a balance between computational efficiency and performance. 
In particular, JiT predicts the clean image in the low-dimensional data manifold, considering the suggestion that high-dimensional or high-frequency noise may distract the model with limited capacity.
Nevertheless, a formidable challenge persists in effectively capturing both intricate image semantics and high-frequency details.
Increasing the model capacity seems to be a straightforward way of tackling this challenge.

Recent advances \cite{ma2025deco, yu2025pixeldit} decouple the generation of high and low frequency components by employing a decoding module to process high-frequency of small patches or pixels.
To mitigate prohibitive memory overhead or OOM issues inherent in small patches, they often resort to removing the attention layer or compressing tokens as a compromise between modeling capacity and computational resources. 
In \cite{ma2025deco}, the low-frequency semantics is up-sampled and re-parameterized as AdaLN layer to modulate the high-frequency of pixel-wise queries, which hinders the propagation of semantics in the global space.
The method \cite{yu2025pixeldit} improves this design by employing global attention to compressed pixel tokens, resulting in lossy information being propagated globally.
We interpret these compromising designs as a mechanism to prioritize low-frequency semantics at the cost of suppression on high-frequency details.

In this paper, we aim to explore the modeling of high and low frequency information in pixel diffusion model without incurring excessive computational cost.
To this end, we design a high-capacity module, called Frequency Booster decoder (FB-Decoder), to enhance the capability of full-frequency modeling.
Specifically, we retain the bottom layers of DiT to generate low-frequency semantics and extract full-frequency information by ``entangling'' the high-frequency queries with low-frequency semantics.
Note that we use the same patch size for DiT and FB-Decoder, thus incurring affordable computational cost even with global attention.
We explore various full-frequency enhancement strategies and propose that simply increasing the feature dimension of FB-Decoder can significantly improve image generation quality.
The negligible effectiveness of the Fusion module in our ablation study indicates the capability of FB-Decoder on full-frequency extraction.

\section{Related Work}
\textbf{Latent Diffusion Models with VAE}. Latent Diffusion Models \cite{rombach2022high, hoogeboom2025simpler} perform the diffusion process within a compact latent space learned by a Variational Autoencoder (VAE). Compared to the raw pixel space, the latent space significantly reduces spatial dimensionality, thereby alleviating both learning complexity and computational overhead. Numerous studies have focused on enhancing the autoencoder by refining architectures or learning objectives, including more robust compression schemes, advanced tokenizers, and in-depth analyses of the optimization trade-offs between reconstruction and generation. However, training a VAE typically involves adversarial objectives and perceptual supervision, which complicates the pipeline. A sub-optimally trained VAE can introduce decoding artifacts, ultimately limiting the synthesis quality of the latent diffusion model. To address this, end-to-end approaches have been explored; for instance, REPA-E\cite{leng2025repa} aligns latent representations with generative objectives by jointly fine-tuning the VAE and the DiT.

\textbf{Representation for Generation}. RAE\cite{zheng2025rae} concurrently generates VAE latent representations and the PCA components of DINOv2\cite{oquab2023dinov2} features within a unified diffusion framework. Similarly, RAE proposes training diffusion models directly on representation encoders (e.g., DINOv2), achieving accelerated convergence. These models strike a balance between high-quality reconstruction and semantically rich latent spaces, while maintaining compatibility with scalable Transformer-based architectures. In a distinct approach, SVG\cite{shi2025latent} also utilizes DINOv2 as an encoder but introduces a trainable Residual Encoder to capture fine-grained details for high-fidelity reconstruction. However, superior reconstruction performance does not inherently translate to enhanced generative capacity; indeed, recent empirical evidence suggests a potential inverse correlation where excessive reconstruction accuracy may undermine synthesis quality. In contrast, Pixel-Space Diffusion Models circumvent the need for reconstruction evaluation by performing image synthesis directly in the pixel domain.

\textbf{Pixel-Space Diffusion Models}. Early research\cite{ho2022cascaded, teng2023relay, li2025fractal,lei2025epg} typically decomposed the diffusion process into multiple resolution stages. However, since computational and memory requirements scale quadratically with image resolution, end-to-end training at the megapixel level remains prohibitively expensive. Relay Diffusion\cite{teng2023relay} addresses this by training independent models for each scale, which inevitably incurs higher costs and necessitates a two-stage optimization. While PixelFlow\cite{chen2025pixelflow} employs a unified model across all scales, it relies on a complex denoising scheduling scheme that compromises inference speed. More recently, PixNerd\cite{wang2025pixnerd} utilizes a DiT to predict patch-wise neural field parameters, achieving pixel-level synthesis efficiency. Meanwhile, works such as DeCo\cite{ma2025deco}, DiP\cite{chen2025dip}, and PixelDiT\cite{yu2025pixeldit} propose incorporating auxiliary pixel decoders to effectively capture challenging high-frequency signals. Empirical evidence from these studies further demonstrates that diminishing the patch size can substantially bolster generative performance. However, as smaller patches significantly increase the memory footprint and the OOM errors, these methods typically involve removing attention layers or employing token compression to strike a pragmatic trade-off between model capacity and computational tractability.
\vspace{-5pt}
\section{Method}
\vspace{-5pt}
In this section, we present FrequencyBooster, an end-to-end Transformer-based diffusion model that performs denoising directly in the pixel space. We begin by providing an overall architecture of FrequencyBooster in Sec.\ref{sec:overall}.  
Subsequently, we detail the proposed FB-Decoder and training loss in Sec.\ref{sec:fb_decoder} and Sec.\ref{sec:loss}, respectively.

\begin{figure*}[h]
    \centering
    \vspace{-10pt}
    \includegraphics[width=0.85\textwidth]{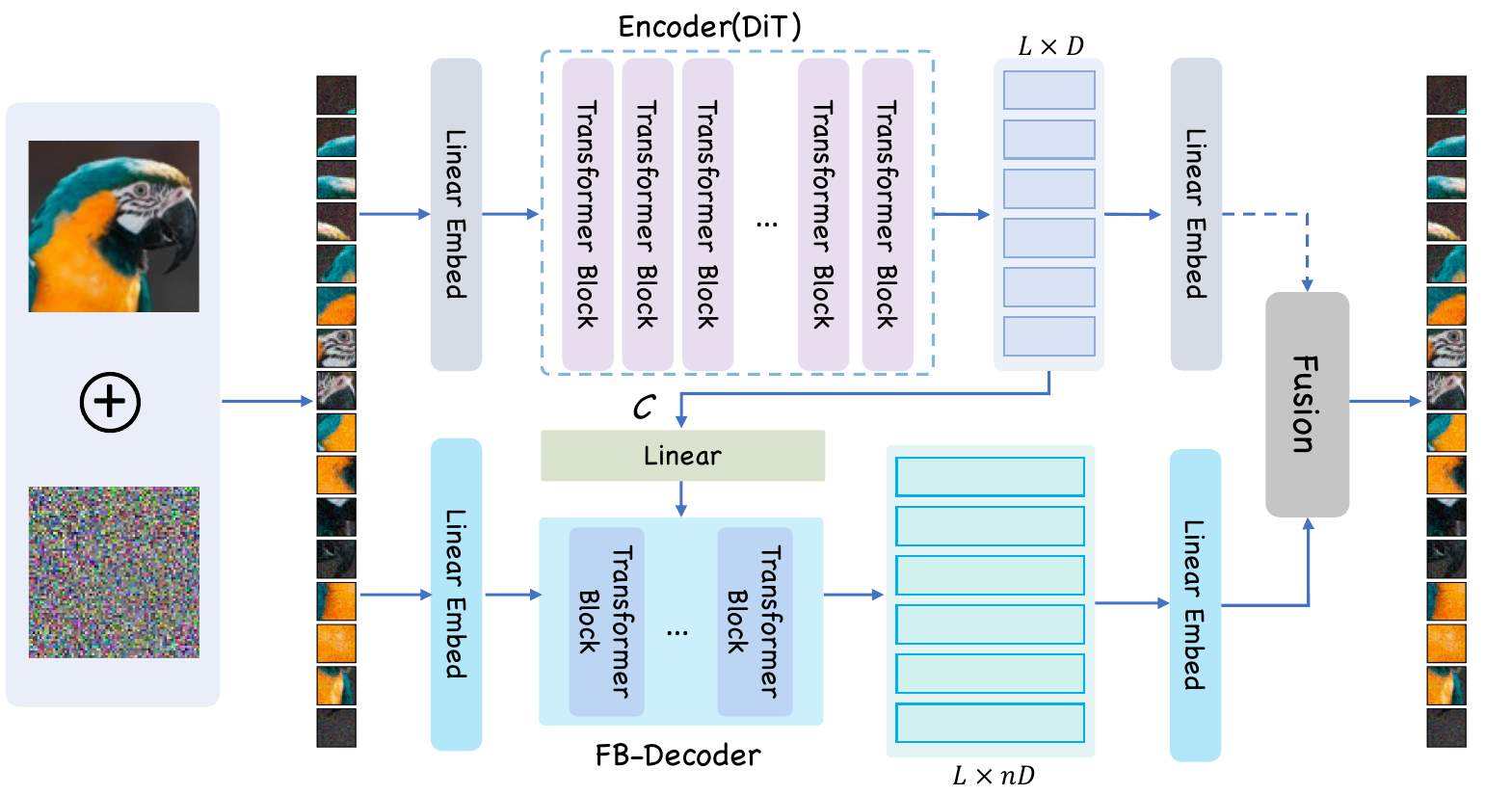}
    \caption{Overview of FrequencyBooster. The architecture integrates a DiT backbone ($L \times D$) and a FB-Decoder ($L \times nD$). An intermediate linear layer bridges the two modules by up-sampling the feature dimensions for high-fidelity pixel generation.}
    \label{fig:main}
    \vspace{-10pt}
\end{figure*}

\subsection{Overall}
\label{sec:overall}
\textbf{Diffusion and Prediction Space.}
Diffusion models can be parameterized in terms of image $x$, noise $\epsilon$, or velocity $v$. Prior methodologies have formulated these outputs as $x$-prediction, $\epsilon$-prediction, and $v$-prediction, respectively.
Recent approaches, such as JiT\cite{li2025jit} and PixelGen\cite{ma2026pixelgen}, have demonstrated significant improvements in $x$-prediction. Following this trajectory, we adopt the image-prediction formulation, specifically targeting $x$-prediction.

We define the data distribution $\mathbf{x} \sim p_{\text{data}}(\mathbf{x})$.
In practice, the noise is typically sampled from a standard multivariate Gaussian distribution, \textit{i.e.}, $\epsilon \sim \mathcal{N}(\mathbf{0}, \mathbf{I})$. 
During the training phase, for a given time step $t \in [0, 1]$, a noisy sample $\mathbf{z}_t$ is defined as:
\begin{equation}
\mathbf{z}_t = t \mathbf{x} + (1 - t) \epsilon .
\end{equation}
The Diffusion Transformer, denoted as $\text{net}_\theta$, predicts the reconstructed image $\mathbf{x}_\theta$ from the noisy sample $\mathbf{x}_t$ as:
\begin{equation}
    \mathbf{x}_\theta = \text{net}_\theta(\mathbf{z}_t, t, \mathbf{c}),
    \label{eq:x0_prediction}
\end{equation}
where $\mathbf{c}$ represents the conditional embedding, typically instantiated as a class label for category-conditional generation.

To preserve the sampling advantages of Flow Matching, we adhere to the methodology of JiT by combining $x$-prediction with $v$-loss. This necessitates converting the predicted image $\mathbf{x}_\theta$ into a velocity field $\mathbf{v}_\theta$, where the predicted velocity $\mathbf{v}_\theta$ is given by:
\begin{equation}
    \mathbf{v}_\theta = \frac{\mathbf{x}_\theta - \mathbf{z}_t}{1 - t}.
\end{equation}

\textbf{The Framework of FrequencyBooster.}  As illustrated in Fig. \ref{fig:main}, the FrequencyBooster framework consists of three pivotal components: the DiT module, the FB-Decoder, and the Fusion module, which collectively orchestrate the image generation and refinement process. 

\begin{figure*}[t]
    \centering
    \vspace{-10pt}
    \begin{minipage}{0.43\textwidth}
        \centering
        \includegraphics[width=\textwidth]{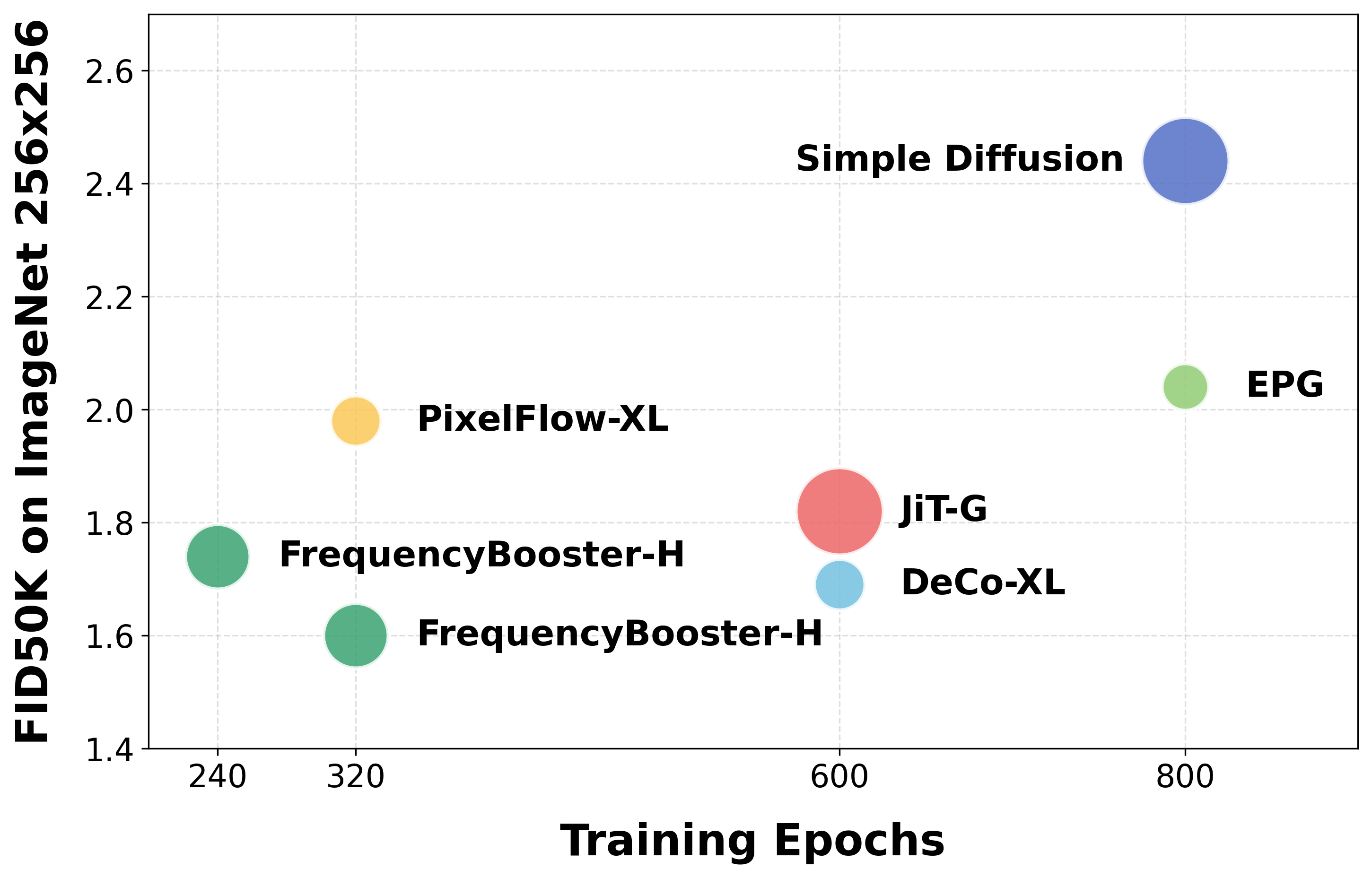}
        \caption{Comparison with other methods. Our method outperforms previous state-of-the-art (SOTA) approaches using only 240 and 320 epochs.}
        \label{fig:bubble}
    \end{minipage}
    \hfill 
    \begin{minipage}{0.5\textwidth}
        \centering
        \includegraphics[width=\textwidth]{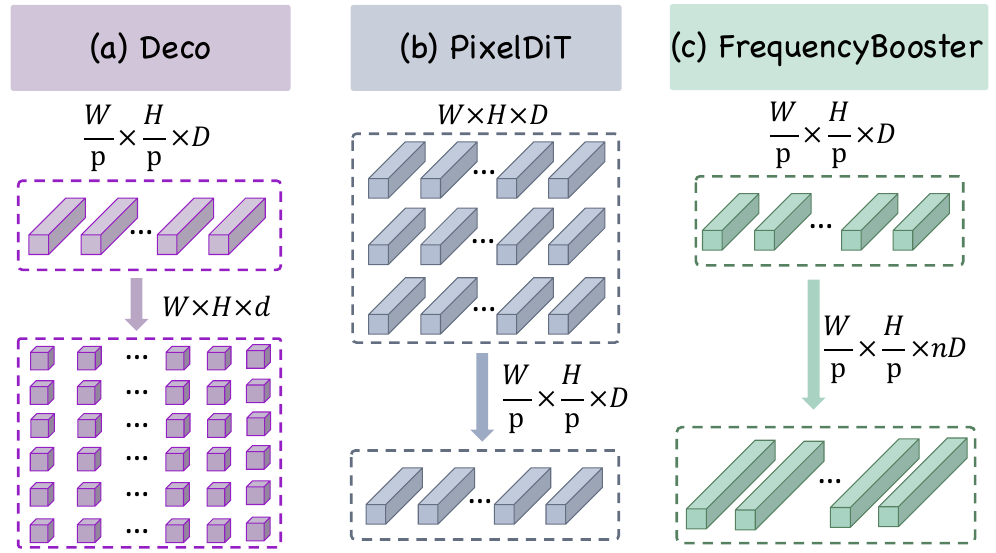}
        \caption{(a) DeCo: $D \to d$ dimensionality reduction ($D \gg d$). (b) PixelDiT: Attention-based token compression. (c) Ours: $D \to nD$ feature expansion.}
        
        \label{fig:feature}
    \end{minipage}
    \vspace{-10pt}
\end{figure*}

While JiT demonstrates the potential of DiT-based $x$-prediction, FrequencyBooster re-purposes the DiT module to specialize in low-frequency semantic modeling. A key distinction lies in our architectural strategy: instead of direct image synthesis, we feed the DiT representations as conditions into a FB-Decoder. 
Our analysis indicates that large-patch yet low-dimensional structures inherently lack the capacity to represent high-frequency information effectively. 
As illustrated in Fig. \ref{fig:feature}, DeCo and PixelDiT attempt to mitigate this ``patch-to-pixel'' gap through degraded decoders by sacrificing architectural complexity or token density. 

FrequencyBooster diverges from this path by leveraging a wider FB-Decoder, which preserves the full information without compromising the expressive power of the attention mechanism.
We also propose a Fusion module to integrate the outputs from the large-patch DiT and the FB-Decoder, facilitating a more effective synthesis of global low-frequency semantics and local high-frequency details.

\subsection{FB-Decoder}
\label{sec:fb_decoder}
As illustrated in Fig. \ref{fig:main}, we propose a dual-stage Transformer architecture. The DiT component is primarily dedicated to capturing coarse-grained structures and low-frequency semantic representations. 
The FB-Decoder is designed to compensate for the high-frequency components and intricate details that are typically vanish in DiT models. Furthermore, we introduce a Fusion module to integrate the outputs from both the DiT and FB-Decoder, leveraging more complementary representations to synthesize high-fidelity images.

Given an input image $\mathbf{x} \in \mathbb{R}^{H \times W \times C}$, we first partition it into non-overlapping patches of size $p \times p$, resulting in a sequence of length $L=(H/p) \times (W/p)$, where each patch is represented as a vector in $\mathbb{R}^{p \times p \times 3}$. These patches are then projected into a latent space via a linear embedding layer, augmented with positional embeddings, and fed into a backbone consisting of stacked Transformer blocks [66]. 
In this way, the semantic condition $\mathbf{c}_s$ is extracted from the DiT module, encapsulating both low-frequency structural information and high-level semantics, as formulated below.

\begin{equation}
    \mathbf{c}_s = \text{DiT}(\mathbf{z}_t, t, \mathbf{c})
\end{equation}
where $t$ denotes the diffusion time-step, and $\mathbf{c}$ represents the class labels or text-based guidance tokens. 

Analogously, the FB-Decoder inputs $\mathbf{z}_t$ and employs a linear layer to re-project the patch representation as $\bar{\mathbf{z}}_t \in \mathbb{R}^{L \times nD}$ where $n \geq 1$. 
Meanwhile, the semantic information $\mathbf{c}_s$ from the DiT module is also fed into the FB-Decoder after dimension up-scaling.
By increasing the feature dimension to $nD$, the FB-Decoder is able to preserve significantly more intricate details and high-frequency information.
Compared with DeCo and PixelDiT, our simple design effectively improves the overall quality and fidelity of the generated image by augmenting the capacity of decoder.
Finally, the FB-Decoder adopts an $x$-prediction formulation as follows:
\begin{equation}
    \mathbf{x}_r = \text{FB-Decoder}(\bar{\mathbf{z}}_t, t, \mathbf{c}_s)
\end{equation}

To further improve the generation fidelity, we introduce a Fusion module to integrate the outputs of DiT and FB-Decoder.
With simple element-wise summation, the Fusion module further supplements the full-frequency information $\mathbf{x}_r$ with global low-frequency semantics $\mathbf{c}_s$.
This architecture facilitates an inherent decoupling of frequency components: the DiT module specializes in capturing coarse-grained structural information, while the FB-Decoder and Fusion module progressively restore intricate textures and fine-grained details with low-frequency semantics, leading to a more natural and balanced learning process.
To generate the pixel image, a re-projection layer is learned to map the representation dimension of fusion tokens into $p \times p \times 3$.

\begin{figure*}[t]
    \centering
    \begin{minipage}{0.46\textwidth}
        \centering
        \includegraphics[width=\textwidth]{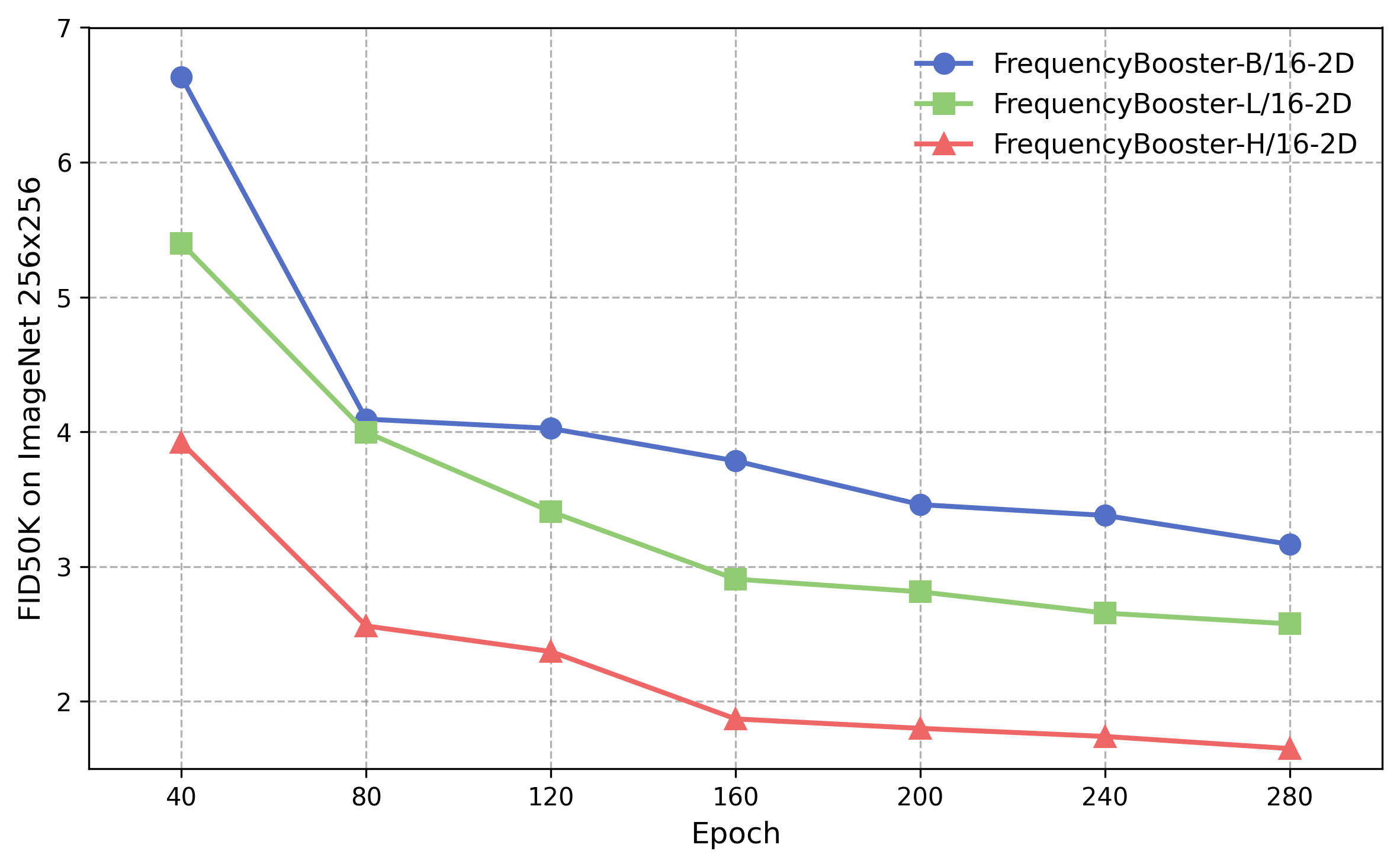}
        \caption{Comparison of gFID scores across model scales (B, L, and XL) on ImageNet $256 \times 256$.}
        \label{fig:model_scale}
    \end{minipage}
    \hfill 
    \begin{minipage}{0.46\textwidth}
        \centering
        \includegraphics[width=\textwidth]{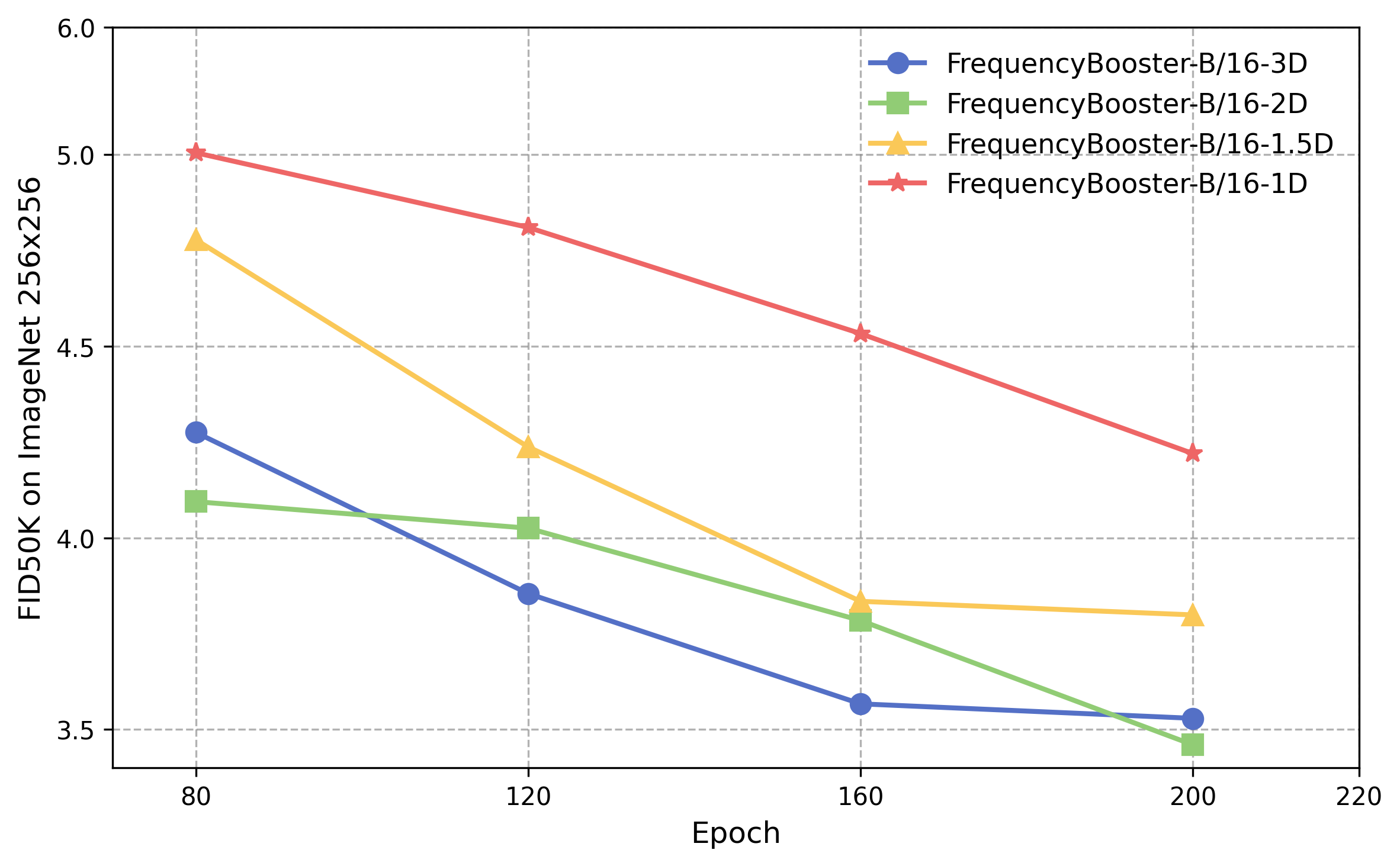}
        \caption{Comparison of gFID scores on ImageNet $256 \times 256$ across various hidden size scales (1D, 1.5D, 2D, and 3D).}
        \label{fig:hidden_scale}
    \end{minipage}
    \vspace{-15pt}
\end{figure*}

\subsection{Training Objectives}
\label{sec:loss}
We adopt the $x$-prediction paradigm optimized via a $v$-loss objective. The $v$-loss is defined as:
\begin{equation}
    \mathcal{L}_{FM} = \mathbb{E}_{t, \mathbf{x}, \epsilon} \left[ \| \mathbf{v}_\theta(\mathbf{z}_t, t) - \mathbf{v} \|^2_2 \right] = \mathbb{E}_{t, \mathbf{x}, \epsilon} \left[ \frac{1}{(1-t)^2} \| \mathbf{x}_\theta - \mathbf{x} \|^2_2 \right]
    \label{fm-x-pred}
\end{equation}

As shown in Eq. (\ref{fm-x-pred}), the $v$-loss can be interpreted as a time-dependent re-weighted formulation of the pixel-space $x$-loss. This formulation harmonizes the inherent stability of $x$-prediction with the superior sampling advantages offered by Flow Matching.

To further enhance the visual fidelity of the synthesized images, we incorporate the LPIPS perceptual loss, which encourages the model to preserve intricate textures and sharp edges that align more closely with human visual perception. Furthermore, to enforce the alignment of intermediate representations and bolster the model's spatial structural awareness, we introduce the iREPA\cite{singh2025irepa}
loss to ensure inherent structural consistency across the dual branches. The total training objective is thus formulated as:
\begin{equation}
\mathcal{L} = \mathcal{L}_{FM}+ \lambda * \mathcal{L}_{iREPA} + \beta * \mathcal{L}_{LPIPS},
\end{equation}
where $\lambda$ and $\beta$ are a hyperparameter used to balance the trade-off between the flow matching loss and the iREPA loss.

\section{Experiments}
\subsection{Implementation Details}
\label{sec 4.1}
We evaluate our approach on the class-conditional ImageNet benchmark. The model is trained with a total batch size of $256$, utilizing DDT as the architectural backbone. Specifically, the hidden dimension of the FB-Decoder is set to $2D$ unless otherwise specified, where $D$ is $768$ and $1024$ in base model and large/huge model, respectively. To improve training stability and model generalization, we maintain an Exponential Moving Average (EMA) of the weights with a decay rate of $0.9999$. Following the protocol in JiT, we incorporate contextual conditioning via the default AdaLN-Zero modulation. Furthermore, inspired by MAR, we prepend a sequence of $32$ specialized tokens to the input; these tokens comprise replicated class embeddings, each integrated with distinct positional embeddings to preserve spatial information.

\begin{figure*}[t] 
    \centering
    \includegraphics[width=0.95\textwidth]{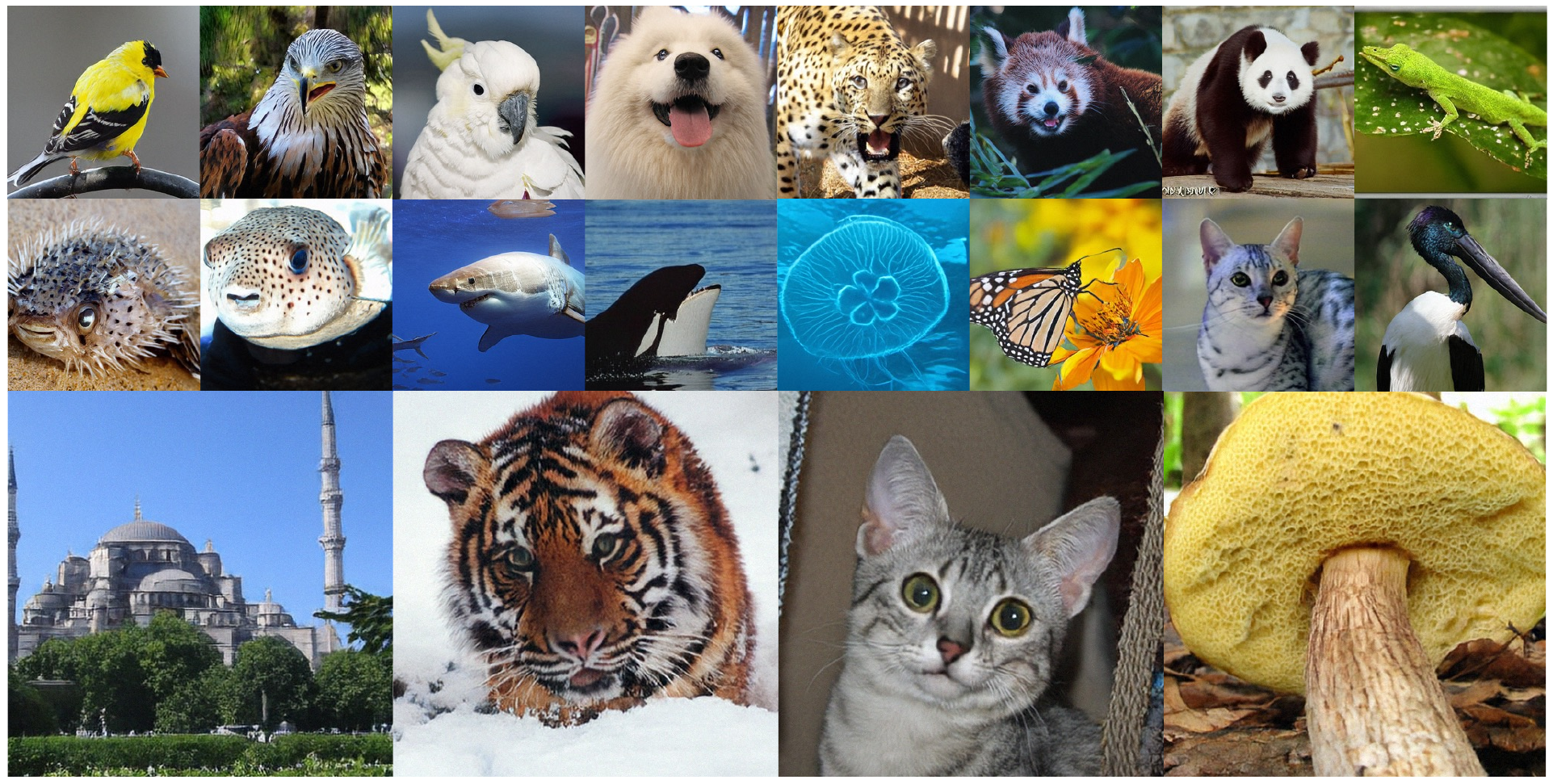}
    \caption{Qualitative results on ImageNet at $256 \times 256$ and $512 \times 512$ resolutions.}
    \label{fig:result}
    \vspace{-10pt}
\end{figure*}

\begin{table*}[!ht]
\centering
\caption{Performance comparison with state-of-the-art models on ImageNet $256 \times 256$ with classifier-free guidance.}
\label{tab:performance}
\footnotesize 
\setlength{\tabcolsep}{4pt} 
\resizebox{\textwidth}{!}{ 
\begin{tabular}{lccccccc} 
\toprule
\multirow{2}{*}{\textbf{Method}} & \multirow{2}{*}{\textbf{Epochs}} & \multirow{2}{*}{\textbf{Params}} & \multirow{2}{*}{\textbf{NFE}} & \multicolumn{4}{c}{\textbf{Generation@256}} \\ \cmidrule(lr){5-8} 
 & & & & \textbf{gFID}$\downarrow$ & \textbf{IS}$\uparrow$ & \textbf{Precision}$\uparrow$ & \textbf{Recall}$\uparrow$ \\ \midrule
\multicolumn{8}{l}{\textit{Latent Generative Models}} \\ \midrule 
LDM-4-G\cite{rombach2022high} & 170 & 400M & - & 3.60 & 247.6 & 0.87 & 0.48 \\
DiT-XL\cite{peebles2023scalable} & 1400 & 675M & $250 \times 2$ & 2.27 & 278.2 & 0.83 & 0.57 \\
SiT-XL\cite{ma2024sit} & 1400 & 675M & $250 \times 2$ & 2.06 & 270.3 & 0.82 & 0.59 \\
MaskDiT\cite{zheng2023fast} & 1600 & 675M & - & 2.28 & 276.5 & 0.80 & 0.61 \\
REPA\cite{yu2025repa} & 800 & 675M & $250 \times 2$ & 1.42 & 305.7 & 0.80 & 0.65 \\
LightningDiT\cite{yao2025reconstruction} & 800 & 675M & - & 1.35 & 295.3 & 0.79 & 0.65 \\
SVG-XL\cite{shi2025latent} & 1400 & 675M & - & 1.92 & 264.9 & - & - \\
DDT-XL\cite{wang2025ddt} & 400 & 675M & - & 1.26 & 310.6 & 0.79 & 0.65 \\
RAE-XL\cite{zheng2025rae} & 800 & 839M & - & \textbf{1.13} & 262.6 & 0.78 & 0.67 \\ \midrule
\multicolumn{8}{l}{\textit{Pixel Generative Models}} \\ \midrule 
StyleGAN-XL\cite{sauer2022stylegan} & - & - & - & 2.30 & 265.1 & 0.78 & 0.53 \\
RIN\cite{jabri2022scalable} & 480 & 410M & - & 3.42 & 182.0 & - & - \\
VDM++\cite{kingma2023understanding} & - & - & $250 \times 2$ & 2.12 & 267.7 & - & - \\
JetFormer\cite{tschannen2024jetformer} & - & 2.8B & - & 6.64 & - & 0.69 & 0.56 \\
Simple Diffusion\cite{hoogeboom2023simple} & 800 & 2.0B & $250 \times 2$ & 2.44 & 256.3 & - & - \\
FARMER\cite{zheng2025farmer} & 320 & 1.9B & - & 3.60 & 269.2 & 0.81 & 0.51 \\
EPG\cite{lei2025epg} & 800 & 583M & - & 2.04 & 283.2 & 0.80 & 0.56 \\
PixelFlow-XL\cite{chen2025pixelflow} & 320 & 677M & $120 \times 2$ & 1.98 & 282.1 & 0.81 & 0.60 \\
PixNerd-XL\cite{wang2025pixnerd} & 320 & 700M & $100 \times 2$ & 1.93 & 298.0 & 0.80 & 0.60 \\
JiT-G/16\cite{li2025jit} & 600 & 2B & $100 \times 2$ & 1.82 & 292.6 & 0.79 & 0.62 \\ 
PixelGen-XL/16\cite{ma2026pixelgen} & 160 & 676M & $100 \times 2$ & 1.83 & 293.6 & 0.79 & 0.63 \\
DeCo-XL/16\cite{ma2025deco} & 320 & 682M & $100 \times 2$ & 1.90 & 303 & 0.80 & 0.61 \\
DeCo-XL/16\cite{ma2025deco} & 600 & 682M & $100 \times 2$ & 1.69 & 304 & 0.79 & 0.63 \\
DiP-XL/16\cite{chen2025dip} & 320 & 631M & $75 \times 2$ & 1.98 & 282.9 & 0.80 & 0.62 \\
DiP-XL/16\cite{chen2025dip} & 600 & 631M & $100 \times 2$ & 1.79 & 281.9 & 0.80 & 0.63 \\ \midrule
\textbf{FrequencyBooster-H/16} & 240 & 1.1B & $100 \times 2$ & 1.74 & \textbf{332} & 0.79 & 0.62 \\
\textbf{FrequencyBooster-H/16} & 320 & 1.1B & $100 \times 2$ & \textbf{1.60} & 321 & 0.80 & 0.63 \\
\bottomrule
\label{table 1}
\end{tabular}
}
\vspace{-30pt}
\end{table*}

For the DiT backbone, we adopt a patch size of $16 \times 16$. The loss weighting coefficients, $\lambda$ and $\beta$, are empirically fixed at $0.05$ and $0.1$, respectively. To bolster generalization and mitigate overfitting, a dropout rate of $0.2$ is consistently applied during training. All experiments are executed on a high-performance cluster equipped with $16$ NVIDIA B200 GPUs.

To rigorously evaluate our model across multiple dimensions, we employ a comprehensive suite of quantitative metrics. We leverage Fréchet Inception Distance (FID) to measure global visual fidelity, while Spatial FID (sFID) is incorporated to further capture structural and spatial coherence. Inception Score (IS) \cite{si2022inception} is utilized to quantify class-conditional diversity. Furthermore, we report Precision and Recall to evaluate sample-level quality and the model's coverage of the data manifold, respectively. Adhering to established protocols, all metrics are computed using $50,000$ generated samples.

\begin{table*}[t]
\centering
\caption{Performance comparison with state-of-the-art models on ImageNet $256 \times 256$ without classifier-free guidance.}
\label{tab:performance}
\footnotesize 
\setlength{\tabcolsep}{3.5pt} 
\resizebox{0.95\textwidth}{!}{ 
\begin{tabular}{lccccccc} 
\toprule
\multirow{2}{*}{\textbf{Method}} & \multirow{2}{*}{\textbf{Epochs}} & \multirow{2}{*}{\textbf{Params}} & \multirow{2}{*}{\textbf{NFE}} & \multicolumn{4}{c}{\textbf{Generation@256}} \\ \cmidrule(lr){5-8} 
 & & & & \textbf{gFID}$\downarrow$ & \textbf{IS}$\uparrow$ & \textbf{Precision}$\uparrow$ & \textbf{Recall}$\uparrow$ \\ \midrule
\multicolumn{8}{l}{\textit{Latent Generative Models}} \\ \midrule 
LDM-4-G\cite{rombach2022high} & 170 & 400M & - & 10.56 & 103.5 & 0.71 & 0.62 \\
DiT-XL\cite{peebles2023scalable} & 1400 & 675M & $250 \times 2$ & 9.62 & 121.5 & 0.67 & 0.67 \\
SiT-XL\cite{ma2024sit} & 1400 & 675M & $250 \times 2$ & 8.61 & 131.7 & 0.68 & 0.67\\
MaskDiT\cite{zheng2023fast} & 1600 & 675M & - & 5.69 & 177.9 & 0.74 & 0.60 \\
REPA-XL-2\cite{yu2025repa} & 800 & 675M & $250 \times 2$ & 5.90 & 157.8 & 0.70 & 0.69 \\
\midrule
\multicolumn{8}{l}{\textit{Pixel Generative Models}} \\ \midrule 
ADM-U\cite{dhariwal2021diffusion}  & 400 & 554M & $250$ & 10.94 & - & 0.69 & 0.63 \\
PixelFlow-XL\cite{chen2025pixelflow} & 320 & 677M & $120 \times 2$ & 12.23 & 103.3 & 0.63 & 0.66 \\
PixNerd-XL\cite{wang2025pixnerd} & 320 & 700M & $100 \times 2$ & 15.61 & 88.9  & 0.59 & 0.68 \\
PixelGen-XL/16\cite{ma2026pixelgen} & 80 & 676M & $100 \times 2$ & 5.11 & 159.2 & 0.72  & 0.63 \\
DeCo-XL/16\cite{ma2025deco} & 320 & 682M & $100 \times 2$ & 14.88 & 88.2  & 0.60  & \textbf{0.68} \\
\textbf{FrequencyBooster-H/16} & 320 & 1.1B & $100 \times 2$ & \textbf{3.79} & \textbf{199.48} & \textbf{0.75} & 0.62 \\

\bottomrule
\label{table 2}
\end{tabular}
}
\vspace{-10pt}
\end{table*}

\begin{table*}[!ht]
\centering
\vspace{-5pt}
\caption{Performance comparison with state-of-the-art models on ImageNet $512 \times 512$ with classifier-free guidance.}
\label{tab:performance}
\footnotesize 
\setlength{\tabcolsep}{4pt} 
\resizebox{\textwidth}{!}{ 
\begin{tabular}{lccccccc} 
\toprule
\multirow{2}{*}{\textbf{Method}} & \multirow{2}{*}{\textbf{Epochs}} & \multirow{2}{*}{\textbf{Params}} & \multirow{2}{*}{\textbf{NFE}} & \multicolumn{4}{c}{\textbf{Generation@512}} \\ \cmidrule(lr){5-8} 
 & & & & \textbf{gFID}$\downarrow$ & \textbf{IS}$\uparrow$ & \textbf{Precision}$\uparrow$ & \textbf{Recall}$\uparrow$ \\ \midrule
\multicolumn{8}{l}{\textit{Latent Generative Models}} \\ \midrule 
DiT-XL\cite{peebles2023scalable} & 600 & 675M & $250 \times 2$ & 3.04 & 240.8 & 0.84 & 0.54 \\
SiT-XL\cite{ma2024sit} & 600 & 675M & $250 \times 2$ & 2.62 & 252.2 & 0.84 & 0.57 \\
\midrule
\multicolumn{8}{l}{\textit{Pixel Generative Models}} \\ \midrule 
ADM-U\cite{dhariwal2021diffusion} & 400 & 554M & $250$ & 7.72 & 172.7 & 0.87 & 0.53 \\
Simple Diffusion\cite{hoogeboom2023simple} & 800 & 2.0B & $250 \times 2$ & 3.54 & 205.0 & - & - \\
PixNerd-XL\cite{wang2025pixnerd} & 340 & 700M & $100 \times 2$ & 2.84 & 245.6 & 0.80 & 0.59 \\
JiT-H/16\cite{li2025jit} & 600 & 965M & $100 \times 2$ & 1.94 & 309.1 & - & - \\ 
DiP-XL/32\cite{chen2025dip}  & - & 631M & $100 \times 2$ & 2.31 & 291.7 & 0.84  & 0.58 \\
DeCo-XL/16\cite{ma2025deco} & 340 & 682M & $100 \times 2$ & 2.22 & 290.0 & 0.80 & 0.60 \\
\midrule
FrequencyBooster-H/16 & 400 & 1.1B & $100 \times 2$ & \textbf{1.69} & \textbf{332.6} & - & - \\
\bottomrule
\label{table 3}
\end{tabular}
}
\vspace{-25pt}
\end{table*}

\subsection{Main Result}
\textbf{Performance on ImageNet $256 \times 256$.}
Table \ref{table 1} presents a comprehensive comparison with recent state-of-the-art (SOTA) methods, all of which employ a Classifier-Free Guidance (CFG)\cite{ho2022classifier} schedule with guidance intervals. With only 320 training epochs and no pre-trained VAE, FrequencyBooster achieves a remarkable FID of 1.60. This significantly outperforms established diffusion models such as DiT-XL (FID 2.27) and SVG-XL (FID 1.92), despite their reliance on substantially longer training schedules. 
FrequencyBooster surpasses the previous state-of-the-art pixel-level model, DeCo-XL/16 (FID 1.69), and significantly outperforms other competitive baselines, including JiT (FID 1.82) and PixelFlow-XL (FID 1.98). Remarkably, even with a truncated training schedule of 240 epochs, our model maintains a competitive FID of 1.74, exceeding well-established models such as DiT-XL that require substantially longer training. 

Table \ref{table 2} reports class-conditional synthesis performance without Classifier-Free Guidance (CFG). Evaluating without CFG provides a direct measure of a model's intrinsic distribution-modeling capacity. In this regime, FrequencyBooster significantly outperforms state-of-the-art methods. With only 320 training epochs, it achieves a 3.79 FID, outperforming latent diffusion models with much longer schedules, such as REPA-XL/2 (5.90 FID at 800 epochs). This advantage underscores that pixel-level diffusion characterizes data distributions more precisely by avoiding the reconstruction artifacts and information loss inherent in VAE-based LDMs. Furthermore, FrequencyBooster yields a $74\%$ and $69\%$ FID improvement over pixel-level counterparts DeCo-XL/16 and PixelFlow-XL, respectively. It also maintains competitive IS, precision, and recall, demonstrating its ability to capture both semantic and structural details. As shown in Fig. \ref{fig:bubble}, our method achieves superior quality and faster convergence than existing approaches.

\textbf{Performance on ImageNet $512 \times 512$.}
Table \ref{table 3} presents a performance comparison on ImageNet $512 \times 512$ using classifier-free guidance. Our FrequencyBooster achieves a state-of-the-art FID of 1.69, significantly outperforming competitive methods such as JiT-H/16 (1.94), DiP-XL/32 (2.31), and DeCo-XL/16 (2.22). Furthermore, our model attains a superior Inception Score (IS) of 332.6, surpassing JiT-H/16 (309.1). These results demonstrate that FrequencyBooster maintains exceptional generation quality and robustness at high resolutions.

\textbf{Qualitative Results.} We present the qualitative results of our model in Fig. \ref{fig:result}. It is evident that our approach excels in synthesizing both global structural integrity and intricate local details, achieving high visual fidelity across diverse samples.


\subsection{Ablation Study}

Tables \ref{tab:architectural} and \ref{tab:loss} present a quantitative evaluation of the architectural components within the proposed FrequencyBooster. By comparing various configurations, we investigate the design choices for the pixel encoder and decoder, the structural internal settings of the decoder, and the strategic integration of perceptual and iREPA losses. All experiments are conducted following the protocols described in Section \ref{sec 4.1}.

\begin{wrapfigure}{r}{0.5\textwidth}
    \centering
    \vspace{-5pt} 
    \includegraphics[width=0.5\textwidth]{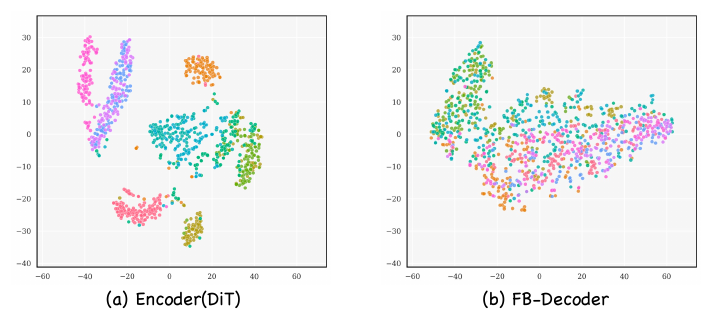}
    \caption{t-SNE visualization of feature spaces. We sampled 100 images per class across 10 ImageNet categories. Features are extracted from our DiT encoder and FB-Decoders; colors denote semantic categories.}
    \vspace{-15pt}
    \label{fig:frequency}
\end{wrapfigure}

\textbf{Qualitative comparison of feature space visualizations via t-SNE.} Fig. \ref{fig:frequency} presents the t-SNE visualization of the feature spaces for the Encoder (DiT) and the FB-Decoder. It is evident that the Encoder (DiT) primarily captures low-frequency semantic information. In contrast, our FB-Decoder supplements the encoder with comprehensive full-frequency features, providing essential fine-grained details.

\begin{table}[t]
  \centering
  \caption{Ablation studies of FrequencyBooster-B/16 on ImageNet 256$\times$256 with Classifier-Free Guidance (CFG). Results start from DiT and incrementally incorporate architectural improvements and loss design strategies.}
  \label{tab:optimized_results}
  
  \begin{subtable}[b]{0.45\textwidth}
    \centering
    \vspace{-5pt}
    \caption{Architectural ablation analysis.}
    \label{tab:architectural}
    \resizebox{\linewidth}{!}{%
    \begin{tabular}{lccc}
      \toprule
      Exp. & Epochs & FID$\downarrow$ & IS$\uparrow$ \\
      \midrule
      $\mathbb{A}$: Baseline-B/16 & 100 & 7.32 & 176.48 \\
      $\mathbb{A}$: Baseline-B/16 & 200 & 4.95 & 233.44 \\
      $\mathbb{B}$: +iREPA & 200 & 3.99 & 263.07 \\
      $\mathbb{C}$: +Wide Decoder & 200 & 3.55 & 287.34 \\
      $\mathbb{D}$: +Fusion & 200 & 3.50 & 296.22 \\
      $\mathbb{E}$: +LPIPS & 200 & 3.40 & 298.42 \\
      \bottomrule
    \end{tabular}%
    
    }
  \end{subtable}
  \hfill
  \begin{subtable}[b]{0.42\textwidth}
    \centering
    \vspace{-5pt}
    \caption{Loss design strategies.}
    
    \label{tab:loss}
    \resizebox{\linewidth}{!}{%
    \begin{tabular}{ccccc}
      \toprule
      Exp. & iREPA & LPIPS & FID$\downarrow$ & IS$\uparrow$ \\
      \midrule
      1 & 0.1  & ---  & 4.08 & 247.33 \\
      2 & 0.05 & ---  & 3.95 & 255.34 \\
      3 & 0.01 & ---  & 4.26 & 250.51 \\
      4 & 0.05 & 0.5  & 3.85 & 247.71 \\
      5 & 0.05 & 0.1  & 3.72 & 258.24 \\
      6 & 0.05 & 0.01 & 3.81 & 256.55 \\
      \bottomrule
    \end{tabular}%
    
    }
  \end{subtable}
  \vspace{-10pt}
\end{table}

\textbf{Contribution of Core Components.} 
Table \ref{tab:architectural} presents an ablation study on the key components of FrequencyBooster-B/16. Starting from a vanilla DiT/16 baseline (FID 4.95 at 200 epochs), the integration of iREPA (Exp. $\mathbb{B}$) successfully drives the FID down to 3.99. By further appending a wide decoder with $2 \times$ hidden dimension (Exp. $\mathbb{C}$), the FID is improved to 3.55. This underscores the superior capacity of the wide decoder over the DiT-only backbone in synthesizing intricate details. Notably, the limited gain observed in Exp. $\mathbb{D}$ which incorporates DiT-layer features—implies that the wide decoder inherently maintains the low-frequency semantics captured by the DiT. The effectiveness of the perceptual loss is further confirmed by comparing Exp. $\mathbb{C}$ and $\mathbb{E}$. We also investigate the sensitivity of loss scaling in Table \ref{tab:loss}, where the configuration of $\lambda=0.05$ (iREPA) and $\beta=0.1$ (LPIPS) yields the most competitive results.

\textbf{Impact of Model Scaling.}  As illustrated in Fig. \ref{fig:model_scale}, we evaluate the performance of FrequencyBooster across three scales—Base (B), Large (L), and Huge (H)—on ImageNet $256 \times 256$ with classifier-free guidance. All variants were trained for 280 epochs. At the end of training, FrequencyBooster-B/16 attains an FID of 3.17, while the FrequencyBooster-L/16 and FrequencyBooster-H/16 variants achieve 2.58 and 1.65, respectively. These results clearly demonstrate that the image generation capability scales significantly with the increase in model capacity.

\section{Conclusions}
We introduced FrequencyBooster, a novel framework that redefines the trajectory of pixel-level generative modeling. Unlike conventional methods that sacrifice spatial details through token compression, our architecture leverages a coupled DiT and FB-Decoder design to maintain a lossless global receptive field while grounding low-frequency semantics. Through our proposed fusion module, we achieve an effective integration of full-frequency information. Notably, by synergizing the iREPA strategy with perceptual loss, we significantly enhance the model's fidelity and training efficiency—attaining SoTA results within a remarkably concise 320-epoch schedule. FrequencyBooster sets a new benchmark across ImageNet datasets, with particularly pronounced advantages in $512 \times 512$ scenarios, highlighting its exceptional scalability. These results validate FrequencyBooster as a powerful and efficient backbone for high-fidelity synthesis. Given its robust performance, our future research will focus on transitioning this architecture to large-scale text-to-image synthesis and fine-grained image editing tasks.

\bibliographystyle{plainnat}
\bibliography{main}

\clearpage 
\appendix

\section{More Implementary Details}
We follow the official implementations of DiT and JiT\cite{li2025jit}. Our settings are outlined in Table \ref{tab:fb_configs}. Further details are provided as follows:

\subsubsection{Detailed Architecture. } 
Distinct from DiT and JiT, our model comprises a DiT backbone and an FB-Decoder. We offer various configurations, including B (Base), L (Large), and H (Huge) variants. Through ablation studies, we investigate the impact of different model scales and the effect of varying the hidden dimension sizes within the FB-Decoder.

\begin{table}[ht]
\centering
\caption{Configurations of experiments.}
\label{tab:fb_configs}
\begin{tabular}{l@{\hspace{5pt}} | @{\hspace{5pt}}c@{\hspace{5pt}}  @{\hspace{5pt}}c@{\hspace{5pt}}  @{\hspace{5pt}}c}
\toprule
 FrequencyBooster& \textbf{Model Size B} & \textbf{Model Size L} & \textbf{Model Size H} \\ \midrule
\rowcolor[HTML]{EFEFEF} 
\textbf{Architecture} & \multicolumn{3}{c}{} \\
DiT depth & 10 & 20 & 28 \\
FB-Decoder depth & 2 & 4 & 4 \\
DiT hidden dim & 768 & 1024 & 1280 \\
FB-Decoder hidden dim & 1536 & 2048 & 2048 \\
heads & 12 & 16 & 16 \\
iREPA loss depth & 4 & 10 & 10 \\
image size & \multicolumn{3}{c}{256 or 512} \\
patch size & \multicolumn{3}{c}{image\_size / 16} \\
bottleneck & \multicolumn{3}{c}{256} \\
dropout & \multicolumn{3}{c}{0.2} \\
in-context class tokens & \multicolumn{3}{c}{32} \\
in-context start block & 4 & 8 & 10 \\ \midrule
\rowcolor[HTML]{EFEFEF} 
\textbf{Training} & \multicolumn{3}{c}{} \\
epochs & \multicolumn{3}{c}{100 or 200 (ablation), 320} \\
optimizer & \multicolumn{3}{c}{Adam, $\beta_1, \beta_2 = 0.9, 0.95$} \\
batch size & \multicolumn{3}{c}{1024} \\
learning rate & \multicolumn{3}{c}{2e-4} \\
learning rate schedule & \multicolumn{3}{c}{constant} \\
weight decay & \multicolumn{3}{c}{0} \\
ema decay & \multicolumn{3}{c}{\{0.9996, 0.9998, 0.9999\}} \\
time sampler & \multicolumn{3}{c}{$\text{logit}(t) \sim \mathcal{N}(\mu, \sigma^2), \mu = -0.8, \sigma = 0.8$} \\
noise scale & \multicolumn{3}{c}{$1.0 \times \text{image\_size} / 256$} \\
clip of $(1-t)$ in division & \multicolumn{3}{c}{0.05} \\
class token drop (for CFG) & \multicolumn{3}{c}{0.1} \\ \midrule
\rowcolor[HTML]{EFEFEF} 
\textbf{Sampling} & \multicolumn{3}{c}{} \\
ODE solver & \multicolumn{3}{c}{Heun} \\
ODE steps & \multicolumn{3}{c}{50} \\
time steps & \multicolumn{3}{c}{linear in [0.0, 1.0]} \\
CFG scale sweep range & \multicolumn{3}{c}{[1.0, 4.0]} \\
CFG interval & \multicolumn{3}{c}{[0.1, 0.95]} \\ \bottomrule
\end{tabular}
\end{table}

\subsubsection{iREPA Loss. } We incorporate the iREPA\cite{singh2025irepa} loss into our framework. Specifically, for the Base variant, the iREPA loss is applied to the 4th layer of the DiT backbone, while for the Large and Huge versions, it is assigned to the 10th layer.

\subsubsection{ImageNet 512 $\times$ 512.} On ImageNet\cite{russakovsky2015imagenet} $512 \times 512$, we adopt the FrequencyBooster/32 configuration, where 32 denotes a patch size of $32 \times 32$. Consistent with the JiT, the model generates 256 tokens for ImageNet $256 \times 256$. The sole distinction lies in the input/output patch dimensionality, which is increased from 768 to 3072 dimensions per patch; otherwise, all other computational processes and overhead remain identical. 

To ensure consistency with the $256 \times 256$ ImageNet baseline, we adjust the noise magnitude for $512 \times 512$ samples by setting $\epsilon \sim \mathcal{N}(0, 2^2 \mathbf{I})$. This adjustment is designed to equilibrate the signal-to-noise ratio (SNR), accounting for the resolution change while keeping the training dynamics invariant.

\subsubsection{In-context class conditioning.} In addition to the standard class conditioning via adaLN-Zero as in DiT, we incorporate in-context tokens following the practice of JiT, with the token count set to 32.

\subsubsection{CFG.} We performed an extensive sweep over the Classifier-Free Guidance (CFG) scale, ranging from 1.0 to 4.0, where a scale of 1.0 denotes the baseline without CFG. Additionally, we investigated the CFG interval and found that the range $[0.1, 0.95]$ yields the optimal results.

\begin{table}[htbp]
    \centering
    \caption{Comparison of performance across different model configurations. Results are reported for FrequencyBooster-B/16 with iREPA loss.}
    \label{ablation_combined}

    \begin{subtable}[b]{0.52\textwidth}
        \centering
        \resizebox{\textwidth}{!}{ 
        \begin{tabular}{c  @{\hspace{5pt}} | @{\hspace{5pt}} c  @{\hspace{5pt}} | @{\hspace{5pt}} c  @{\hspace{5pt}} | @{\hspace{5pt}} c @{\hspace{5pt}} | @{\hspace{5pt}} c @{\hspace{5pt}} | @{\hspace{5pt}} c }
            \toprule
            Exp. & \makecell{Encoder\\Layers} & \makecell{Decoder\\Layers} & \makecell{Hidden\\Size} & FID $\downarrow$ & IS $\uparrow$ \\
            \midrule
            1 & 12 & 0& 768 & 4.64 & 233.40 \\
            2 & 10 & 2& 768 & 4.95 & 233.44 \\
            3 & 10 & 2& $768 \times 2 $   & 4.05 & 278.33 \\
            \bottomrule
        \end{tabular}
        }
        \subcaption{Impact of hidden size. (FrequencyBooster-B/16, trained for 200 epochs without auxiliary loss.)}
        \label{tab:hidden}
    \end{subtable}
    \hfill 
    \begin{subtable}[b]{0.41\textwidth}
        \centering
        \resizebox{\textwidth}{!}{ 
        \begin{tabular}{c @{\hspace{5pt}} | @{\hspace{5pt}} c  @{\hspace{5pt}} | @{\hspace{5pt}} c  @{\hspace{5pt}} | @{\hspace{5pt}} c  @{\hspace{5pt}} | @{\hspace{5pt}} c}
            \toprule
            Exp. & \makecell{Encoder\\Layers} & \makecell{Decoder\\Layers} & FID $\downarrow$ & IS $\uparrow$ \\
            \midrule
            1 & 10 & 2 & \textbf{5.20} & 232.70 \\
            2 & 8  & 4 & 5.35 & 225.18 \\
            3 & 6  & 6 & 6.61 & 186.12 \\
            \bottomrule
        \end{tabular}
        }
        \subcaption{Impact of layer depth (100 epochs). FrequencyBooster-B/16 with iREPA loss.}
        \label{tab:layers}
    \end{subtable}
\end{table}

\section{Additional Ablation Studies}
\subsubsection{Ablation Studies of CFG.} As illustrated in Figure \ref{fig:cfg_ablation}, we demonstrate the impact of CFG settings on the performance of FrequencyBooster-H trained for 240 epochs on the ImageNet $256 \times 256$ dataset. Furthermore, Table \ref{tab:cfg_intervals} indicates that at 320 training epochs, FrequencyBooster-H achieves optimal results when the CFG scale is set to 3.0 and the CFG interval is [0.1, 0.95].

\subsubsection{Exploration of Encoder-Decoder Depth.} 
As illustrated in Table \ref{tab:hidden}, a comparison across Exp. 1 $\sim$ 3 for the FrequencyBooster-B/16 variant reveals that the configuration with 10 encoder layers and 2 decoder layers achieves the best performance.

\subsubsection{Impact of Hidden Representation Capacity.}  As illustrated in Table \ref{tab:layers}, we evaluate the FrequencyBooster-B/16 variant without auxiliary losses. A comparison between Exp. 1 and Exp. 2 reveals that given the same hidden size and total layer count, the 10-encoder/2-decoder architecture underperforms the vanilla 12-layer DiT, suggesting that the asymmetric encoder-decoder structure does not provide inherent architectural gains. However, comparing Exp. 2 and Exp. 3 demonstrates that doubling the hidden size leads to a substantial performance boost. This indicates that a decoder with a larger hidden dimension significantly enhances the model's capacity for capturing image semantics and fine-grained structural details. Fig. \ref{fig:hidden_scale} illustrates the performance across various hidden sizes ($1D, 1.5D, 2D, \text{and } 3D$), where $D$ denotes the base hidden dimension. Our experiments reveal that the hidden size significantly influences model capacity. Notably, the performance gap between $2D$ and $3D$ configurations is marginal, suggesting that the model capacity is approaching a saturation point.

\subsubsection{Ablation Studies of patch sizes and hidden sizes.} As illustrated in Table \ref{tab:comparison}, we evaluate FrequencyBooster-B on ImageNet $256 \times 256$ (200 epochs). A configuration with patch size 16 and hidden size 768 yields an FID of 4.22. Reducing the patch size to 8 quadruples the token count and improves the FID to 4.01, confirming that finer patch granularity enhances generation quality at the cost of significantly higher computational overhead. However, a patch size of 4 triggers Out-of-Memory (OOM) errors, indicating that escalating VRAM requirements pose a substantial bottleneck. In contrast, our further experiments demonstrate that increasing the hidden size while keeping the patch size constant also markedly boosts performance without encountering OOM issues, offering a more hardware-friendly scaling path.

\begin{figure*}[!ht]
    \centering
    \begin{minipage}[c]{0.48\linewidth}
        \centering
        \captionof{table}{Ablation study of CFG and CFG intervals on FID and IS for FrequencyBooster-H.}
        \label{tab:cfg_intervals}
        \footnotesize 
        \begin{tabular*}{\linewidth}{@{\extracolsep{\fill}}cccc@{}} 
            \toprule
            CFG ($w$) & CFG Interval & FID $\downarrow$ & IS $\uparrow$ \\ 
            \midrule
            2.9 & [0.1, 1.0] & 1.631 & 319.04 \\ 
            3.0 & [0.1, 1.0] & 1.630 & 321.36 \\ 
            3.1 & [0.1, 1.0] & 1.622 & 319.39 \\ 
            \textbf{3.1} & \textbf{[0.1, 0.95]} & \textbf{1.602} & \textbf{324.48} \\ 
            3.1 & [0.1, 0.90] & 1.616 & 319.39 \\ 
            \bottomrule
        \end{tabular*}
    \end{minipage}
    \hspace{0.02\linewidth} 
    \begin{minipage}[c]{0.48\linewidth}
        \centering
        \captionof{table}{ Comparison of different patch size and hidden sizes (ImageNet $256 \times 256$).}
        \label{tab:comparison}
        \footnotesize 
        \begin{tabular*}{\linewidth}{@{\extracolsep{\fill}}cccc@{}} 
            \toprule
            Patch Size & Hidden Size & FID $\downarrow$ & IS $\uparrow$ \\ 
            \midrule
            16 & 768 & 4.22 & 325.90 \\ 
            8 & 768 & 4.01 & 328.81 \\
            16 & 768 $\times$ 1.5 & 3.80 & 313.76 \\
            16 & 768 $\times$ 2 & 3.46 & 314.25 \\
            16 & 768 $\times$ 3 & 3.53 & 315.22 \\
        
            \bottomrule
        \end{tabular*}
    \end{minipage}
\end{figure*}

\section{Spectral Analysis.} 

We analyze the spectral characteristics of the encoder (DiT) and FB-Decoder modules in our method compared to JiT. Specifically, we apply the 2D Discrete Fourier Transform (DFT) to the spatial feature maps and compute their channel-averaged magnitude spectra.
\begin{figure*}[!ht]
    \centering
    \begin{minipage}[c]{0.47\linewidth}
        \centering
        \includegraphics[width=\linewidth]{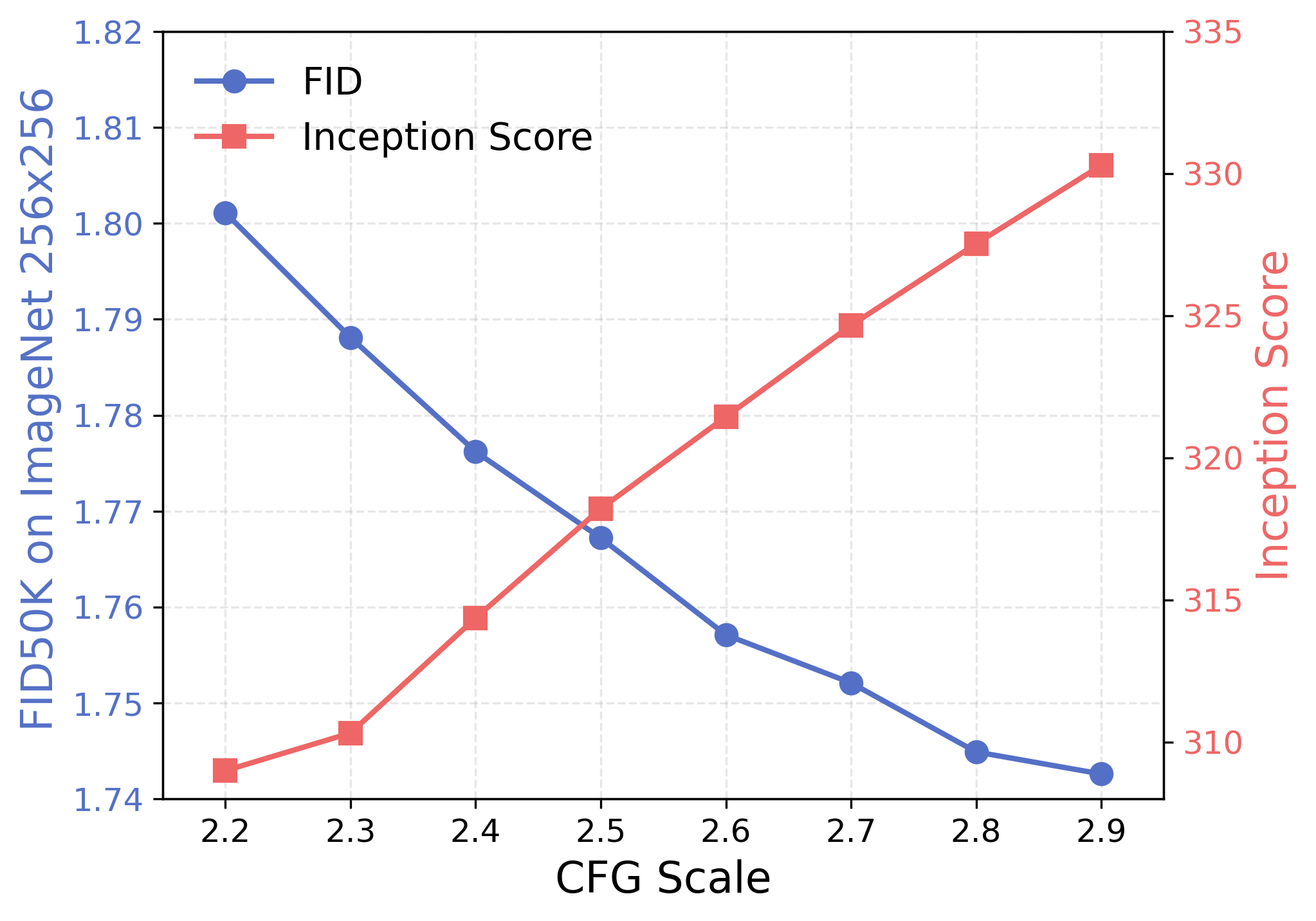}
        \caption{Ablation study of the CFG scale. We evaluate the generation quality of FrequencyBooster-H using FID ($\downarrow$) and IS ($\uparrow$) on ImageNet $256 \times 256$ after 240 training epochs} 
        \label{fig:cfg_ablation}
    \end{minipage}
    \hspace{0.04\linewidth} 
    %
    \begin{minipage}[c]{0.47\linewidth}
        \centering
        \includegraphics[width=\linewidth]{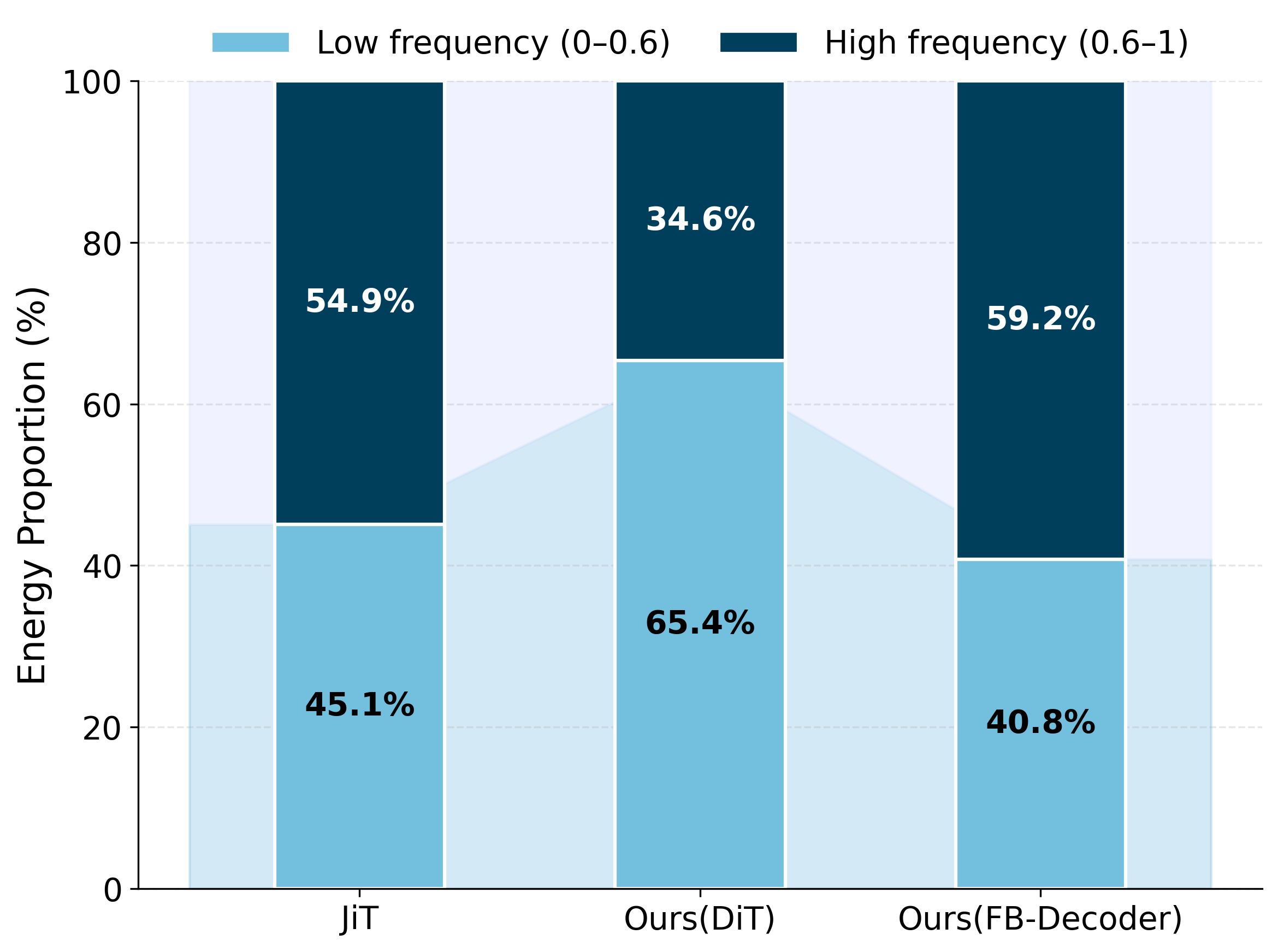}
        \caption{Spectral Analysis: Radial frequencies are normalized to $f \in [0, 1]$, with energy ratios calculated for low-frequency ($f \in [0, 0.6]$) and high-frequency ($f \in (0.6, 1.0]$) regions.} 
        \label{fig:spectral_analysis}
    \end{minipage}
\end{figure*}

 As illustrated in Figure \ref{fig:spectral_analysis}, our encoder primarily focuses on low-frequency semantic information. In contrast, the FB-Decoder significantly replenishes the high-frequency components missing in the encoder. Compared to JiT, our FB-Decoder exhibits richer high-frequency content, enabling the extraction of both intricate details and global semantics. This demonstrates that our model empowers the pixel diffusion process with comprehensive full-frequency modeling capabilities.

\begin{figure*}[h]
    \centering
    \includegraphics[width=0.98\textwidth]{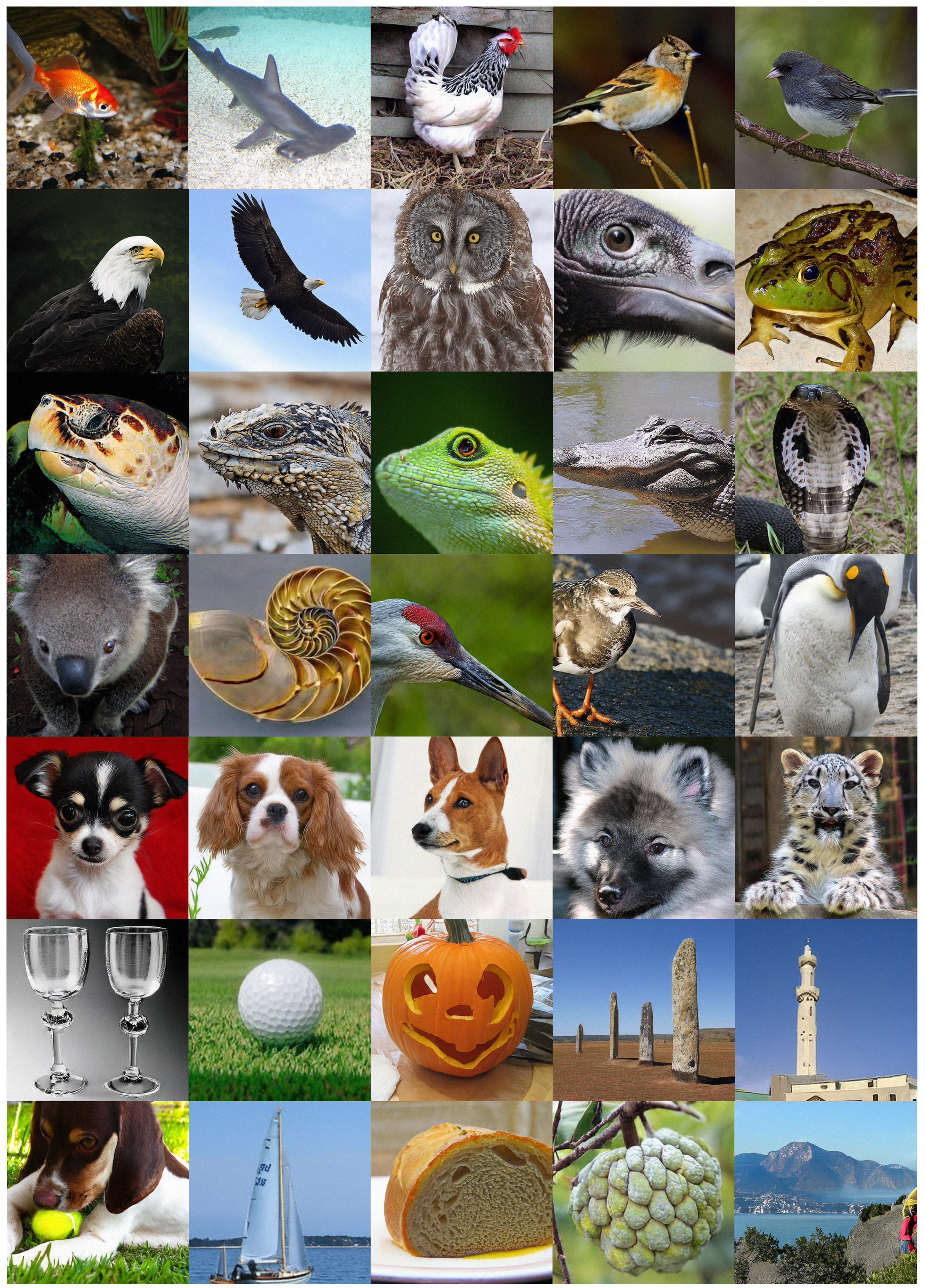}
    \caption{Qualitative result on the ImageNet dataset at $512 \times 512$ resolution.}
    \label{fig:main}
\end{figure*}

\begin{figure*}[h]
    \centering
    \includegraphics[width=0.98\textwidth]{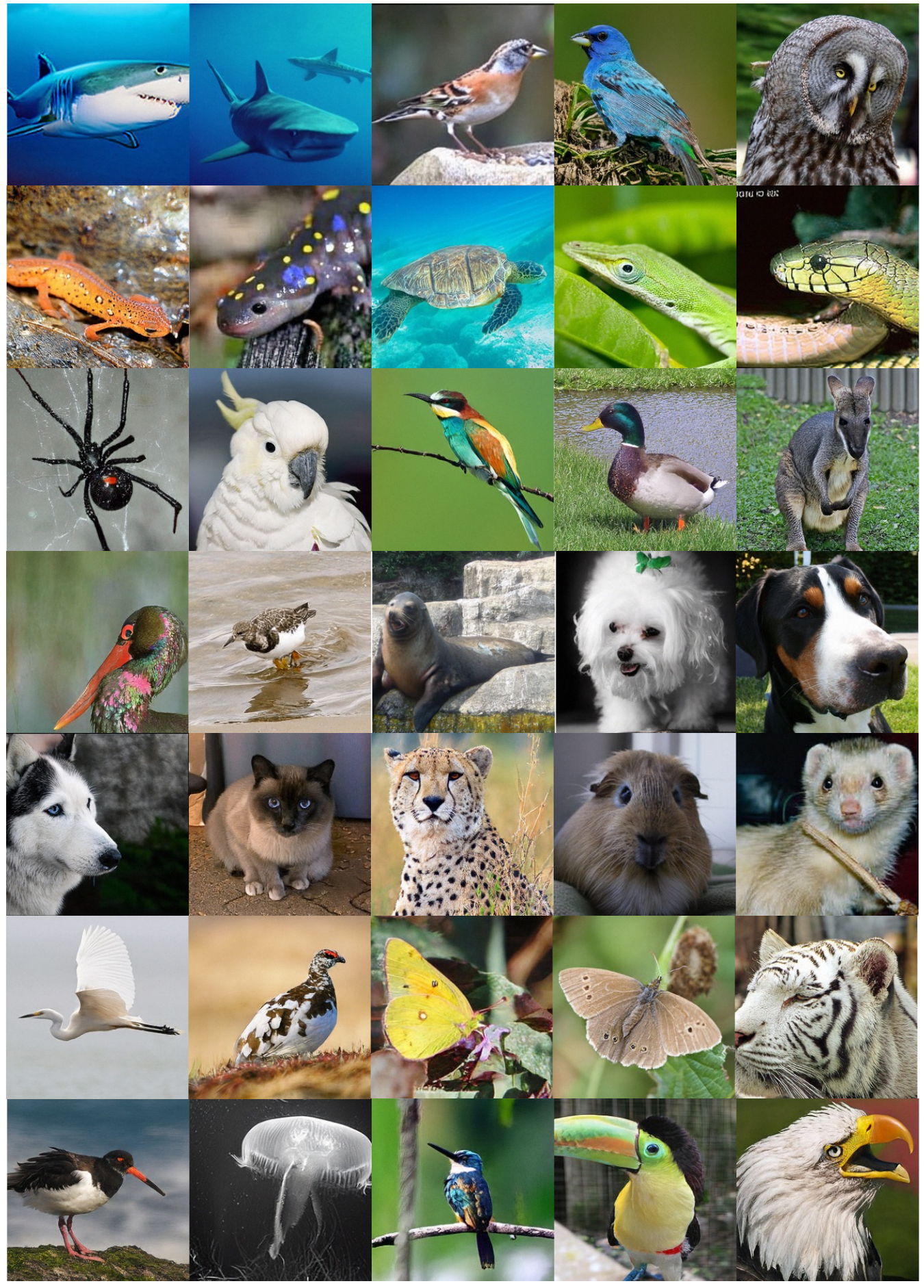}
    \caption{Qualitative result on the ImageNet dataset at $256 \times 256$ resolution.}
    \label{fig:main}
\end{figure*}

\clearpage  

\end{document}


\maketitle

\appendix

\section{Technical appendices and supplementary material}
Technical appendices with additional results, figures, graphs, and proofs may be submitted with the paper submission before the full submission deadline (see above). You can upload a ZIP file for videos or code, but do not upload a separate PDF file for the appendix. There is no page limit for the technical appendices. 

Note: Think of the appendix as ``optional reading'' for reviewers. The paper must be able to stand alone without the appendix; for example, adding critical experiments that support the main claims to an appendix is inappropriate. 


\newpage
\section*{NeurIPS Paper Checklist}

\begin{enumerate}

\item {\bf Claims}
    \item[] Question: Do the main claims made in the abstract and introduction accurately reflect the paper's contributions and scope?
    \item[] Answer: \answerYes{} 
    \item[] Justification: The abstract and introduction outline our theoretical and technical contributions, the proposed methodology, and a comprehensive experimental evaluation.

\item {\bf Limitations}
    \item[] Question: Does the paper discuss the limitations of the work performed by the authors?
    \item[] Answer: \answerYes{} 
    \item[] Justification: At the end of the conclusion section, a limitation discussion is provided. 

\item {\bf Theory assumptions and proofs}
    \item[] Question: For each theoretical result, does the paper provide the full set of assumptions and a complete (and correct) proof?
    \item[] Answer: \answerYes{} 
    \item[] Justification: Yes, the assumptions are given in the form of Lema in the Prelimaries section. A complete proof is given in the method section.

    \item {\bf Experimental result reproducibility}
    \item[] Question: Does the paper fully disclose all the information needed to reproduce the main experimental results of the paper to the extent that it affects the main claims and/or conclusions of the paper (regardless of whether the code and data are provided or not)?
    \item[] Answer: \answerYes{} 
    \item[] Justification: Yes, the baseline model, technical details, hyperparameters, and configuration are detailed in the submission for reproducibility.

\item {\bf Open access to data and code}
    \item[] Question: Does the paper provide open access to the data and code, with sufficient instructions to faithfully reproduce the main experimental results, as described in supplemental material?
    \item[] Answer: \answerYes{} 
    \item[] Justification: The datasets this paper uses are publicly available, and the source code is promised to be public once published.

\item {\bf Experimental setting/details}
    \item[] Question: Does the paper specify all the training and test details (e.g., data splits, hyperparameters, how they were chosen, type of optimizer) necessary to understand the results?
    \item[] Answer: \answerYes{} 
    \item[] Justification: The implementation details are given at the beginning of the experimental section and the beginning of the appendix.

\item {\bf Experiment statistical significance}
    \item[] Question: Does the paper report error bars suitably and correctly defined or other appropriate information about the statistical significance of the experiments?
    \item[] Answer: \answerNo{} 
    \item[] Justification: Following prior works in this field, the evaluation protocols on the corresponding datasets do NOT require a report of the error bar.

\item {\bf Experiments compute resources}
    \item[] Question: For each experiment, does the paper provide sufficient information on the computer resources (type of compute workers, memory, time of execution) needed to reproduce the experiments?
    \item[] Answer: \answerYes{} 
    \item[] Justification: The implementation details are given at the beginning of the experimental section and the beginning of the appendix.
    
\item {\bf Code of ethics}
    \item[] Question: Does the research conducted in the paper conform, in every respect, with the NeurIPS Code of Ethics \url{https://neurips.cc/public/EthicsGuidelines}?
    \item[] Answer: \answerYes{} 
    \item[] Justification: This paper focuses on a fundamental task of machine learning and conducts experiments on publicly available datasets.

\item {\bf Broader impacts}
    \item[] Question: Does the paper discuss both potential positive societal impacts and negative societal impacts of the work performed?
    \item[] Answer: \answerYes{} 
    \item[] Justification: The societal impact of this work has been discussed at the end of the conclusion section. We do not envision negative societal impact could be brought by this work.
    
\item {\bf Safeguards}
    \item[] Question: Does the paper describe safeguards that have been put in place for responsible release of data or models that have a high risk for misuse (e.g., pre-trained language models, image generators, or scraped datasets)?
    \item[] Answer: \answerNA{} 
    \item[] Justification: This work focuses on a fundamental problem in pixel generation and conducts experiments on standard datasets. We do not envision such risks.

\item {\bf Licenses for existing assets}
    \item[] Question: Are the creators or original owners of assets (e.g., code, data, models), used in the paper, properly credited and are the license and terms of use explicitly mentioned and properly respected?
    \item[] Answer: \answerYes{} 
    \item[] Justification: All the assets have been properly cited, with a license to use for academia and no commercial purpose.

\item {\bf New assets}
    \item[] Question: Are new assets introduced in the paper well documented and is the documentation provided alongside the assets?
    \item[] Answer: \answerNA{} 
    \item[] Justification: This paper does not release new assets.

\item {\bf Crowdsourcing and research with human subjects}
    \item[] Question: For crowdsourcing experiments and research with human subjects, does the paper include the full text of instructions given to participants and screenshots, if applicable, as well as details about compensation (if any)? 
    \item[] Answer: \answerNA{} 
    \item[] Justification: This paper does not involve crowdsourcing nor research with human subjects.

\item {\bf Institutional review board (IRB) approvals or equivalent for research with human subjects}
    \item[] Question: Does the paper describe potential risks incurred by study participants, whether such risks were disclosed to the subjects, and whether Institutional Review Board (IRB) approvals (or an equivalent approval/review based on the requirements of your country or institution) were obtained?
    \item[] Answer: \answerNA{} 
    \item[] Justification: This paper does not involve crowdsourcing nor research with human subjects

\item {\bf Declaration of LLM usage}
    \item[] Question: Does the paper describe the usage of LLMs if it is an important, original, or non-standard component of the core methods in this research? Note that if the LLM is used only for writing, editing, or formatting purposes and does \emph{not} impact the core methodology, scientific rigor, or originality of the research, declaration is not required.
    \item[] Answer: \answerNA{} 
    \item[] Justification: The core method development in this research of this paper does not involve LLMs as any important, original, or non-standard components.

\end{enumerate}